\documentclass[sigconf]{acmart}
\usepackage{multirow}
\usepackage{booktabs}
\usepackage{subfigure}

\usepackage[normalem]{ulem}
\useunder{\uline}{\ul}{}
\AtBeginDocument{%
  \providecommand\BibTeX{{%
    \normalfont B\kern-0.5em{\scshape i\kern-0.25em b}\kern-0.8em\TeX}}}

\setcopyright{acmcopyright}
\copyrightyear{2022}
\acmYear{2022}
\acmDOI{XXXXXXX.XXXXXXX}

\acmConference[KDD-UC ’22]{August 14–18, 2022 }{August 14–18, 2022}{Washington, DC, USA}
\acmPrice{15.00}
\acmISBN{978-1-4503-XXXX-X/18/06}

\begin{document}

\title{Knowledge Graph Completion with Pre-trained Multimodal Transformer and Twins Negative Sampling}

\author{Yichi Zhang}
\email{3180103772@zju.edu.cn}
\orcid{1234-5678-9012}
\affiliation{%
  \institution{College of Computer Science and Technology, Zhejiang University}
  \city{Hangzhou, Zhejiang}
  \country{China}
}

\author{Wen Zhang}
\authornote{Corresponding Author.}
\email{wenzhang2015@zju.edu.cn}
\affiliation{%
  \institution{College of Software Technology, Zhejiang University}
  \city{Hangzhou, Zhejiang}
  \country{China}
 }


\renewcommand{\shortauthors}{Yichi Zhang et al.}

\begin{abstract}
Knowledge graphs (KGs) that modeling the world knowledge as structural triples are inevitably incomplete. Such problems still exist for multimodal knowledge graphs (MMKGs). Thus, knowledge graph completion (KGC) is of great importance to predict the missing triples in the existing KGs. As for the existing KGC methods, embedding-based methods rely on manual design to leverage multimodal information while finetune-based approaches are not superior to embedding-based methods in link prediction. To address these problems, we propose a \textbf{V}isual\textbf{B}ERT-enhanced \textbf{K}nowledge \textbf{G}raph \textbf{C}ompletion model (VBKGC for short). VBKGC could capture deeply fused multimodal information for entities and integrate them into the KGC model. Besides, we achieve the co-design of the KGC model and negative sampling by designing a new negative sampling strategy called twins negative sampling. Twins negative sampling is suitable for multimodal scenarios and could align different embeddings for entities. We conduct extensive experiments to show the outstanding performance of VBKGC on the link prediction task and make further exploration of VBKGC.

\end{abstract}




\maketitle

\section{Introduction}
Knowledge graphs (KGs) represent world knowledge as structured triples in the form of (head entity, relation, tail entity), $(h, r, t)$ for short, which means entity $h$ and $t$ have a relation $r$. KGs contain a wealth of structural information of world knowledge and have become the infrastructure of AI research and can benefit a lot of tasks like recommendation systems \cite{zhang2021billion}, language modeling \cite{liu2020k}  and question answering \cite{yasunaga2021qa}.
\par Multimodal knowledge graphs (MMKGs) \cite{liu2019mmkg} are KGs containing a wealth of modal information (images and text), which greatly enhances the expressiveness of the KGs. How to leverage the modal information in multimodal knowledge graphs is a current research hotspot in KG-related research.
\par However, KGs and of course MMKGs are far from complete. They contain only the knowledge we have observed while plenty of triples are not discovered among the entities and relations. Therefore, knowledge graph completion (KGC) is an important task in KG research which aims to discover the missing triples in KGs. As for MMKGs, the approaches to achieve multimodal knowledge graph completion (MMKGC) can be divided into two main categories: embedding-based approaches \cite{xie2016image, mousselly2018multimodal, pezeshkpour2018embedding} and finetune-based approaches. Embedding-based approaches follow the paradigm of knowledge graph embedding (KGE) which embed the entities and relations into a low-dimensional vector space and define a score function to estimate the plausibility of triples. We could call the embedding-based approaches multimodal knowledge graph embedding (MMKGE) as multimodal information in the KGs is also considered in the embedding model. finetune-based approaches \cite{yao2019kg, kim2020multi, wang2021structure} usually employ pre-trained language models like BERT \cite{devlin2018bert} and encode the triples with their textual descriptions. Then the models are finetuned with the KG-related tasks like triple classification \cite{yao2019kg} and relation prediction \cite{kim2020multi}. The triple scores are based on the output of the BERT model.
\par Although existing methods mentioned above have made strides in MMKGC, these methods face the following problems. (1) For the embedding-based methods, their ability to utilize multimodal information about entities is insufficient. The extraction and fusion of multimodal information are highly dependent on the manual design and need more work to find the most suitable method for each dataset. For example, \cite{wang2019multimodal} propose three different methods to achieve modal fusion and search for the best strategy on two datasets. (2) For the finetune-based approaches, though they leverage the textual information by fine-tuning the pre-trained model, the inference speed on the test set is unbearably slow \cite{wang2021structure} due to the deep architecture of the pre-trained model and rank-based evaluation protocol for KGs. The evaluation results of them are still not significantly better than embedding-based methods either. (3) Design of negative sampling is ignored in all of the approaches. Existing methods just apply the normal negative sampling for model training, which might not be suitable for the multimodal scenario. For the multiple embeddings in the multimodal scenario, aligning them is also important. Normal negative sampling is entity-level and has no such ability.
\par To address the problems mentioned above, we propose a multimodal knowledge graph completion model called \textbf{V}isual\textbf{B}ERT-enhanced \textbf{K}nowledge \textbf{G}raph {C}ompletion model (VBKGC for short). VBKGC is an embedding-based model which employs a pre-trained multimodal transformer model (VisualBERT \cite{li2019visualbert} for example) to extract deeply fused multimodal feature which is free of finetuning and 
have a fast inference speed like many other embedding-based methods. Besides, we achieve the co-design of the MMKGE model and negative sample strategy. We propose a negative sample strategy called twins negative sampling for MMKGE. Twins negative sampling could align the different embeddings of each entity during training and achieve better performance on link prediction tasks.
\par In general, our contributions in this paper can be summarized as follows:
\begin{itemize}
    \item We propose an MMKGC model called VBKGC, which is an embedding-based model and employs VisualBERT as a multimodal encoder to capture the deeply fused multimodal features of entities. It is a universal approach and needs no more manual design for modal feature extraction and fusion. Besides, VBKGC has fast inference speed.
    \item We achieve the co-design of the model and negative sampling for KGC. We propose a new negative sampling method called twins negative sampling for multimodal scenarios. Twins negative sampling could align the structural and multimodal embeddings for entities to perform better on the link prediction task.
    \item We conduct comprehensive experiments on link prediction tasks with two benchmark datasets. We make further exploration of VBKGC model and twins negative sampling.
\end{itemize}

\section{Related Works}

\subsection{Knowledge Graph Embedding}
Knowledge graph embedding (KGE) \cite{wang2017knowledge} aims to embed entities and relations of KGs into a low-dimensional continuous vector space and measure the plausibility of triples by a well-defined score function, which is a popular topic in KG-related research. 
\par Existing KGE models are diversified. Translation-based methods like TransE \cite{transe} and TransH \cite{transh} modeling the relation in each triple as a translation from head entity to tail entity. Sematic-based methods like DistMult \cite{distmult} and ComplEx \cite{trouillon2017knowledge} apply similarity-based score function to modeling the triples. Other method such as RotatE \cite{rotate} and ConE \cite{zhang2021cone} also modeling triples with various mathematical structures. Convolutional neural networks and graph neural networks are also employed in some KGE models \cite{dettmers2018convolutional, jiang2019adaptive, vashishth2019composition, schlichtkrull2018modeling}, which play a role as feature encoders. Rule-enhanced methods \cite{zhang2019iteratively, zhang2019interaction} integrate rule learning in KGE for better performance and explainability.
\par Meanwhile, negative sampling (NS) \cite{transe, transh} is a key technology for KGE. It would generate negative triples and teach the KGE model to distinguish between positive and negative triples. Many researchers propose better negative sampling strategies. GAN-based methods \cite{cai2017kbgan, wang-etal-2020-incorporating} apply GAN \cite{goodfellow2014generative} to generate hard negative triples. NSCaching \cite{zhang2019nscaching} simply stores the high-quality negative triples with cache during training. Other methods like SANS \cite{ahrabian2020structure} and CAKE \cite{niu2022cake} leverage information from original KGs and sample high-quality negative triples.
\par However, existing NS methods are usually designed for general KGE. In this context, general KGE means KGE with no extra information outside the triplet structure. In the paragraphs that follow, we still use such a concept. In the multimodal scenario, each entity in a KG might have multiple embeddings (structural embedding and multimodal embedding for example) rather than only one structural embedding, which means the aligning the multiple embeddings is also of great significance. Unfortunately, the existing NS methods do not have this feature and a new NS method urgently needs to be proposed.

\subsection{Multimodal Knowledge Graph Completion}
Multimodal knowledge graph completion (MMKGC) is an important task for MMKGs, which would predict the missing triples in MMKGs with multimodal information of entities and relations. Previous methods of MMKG could be roughly divided into two categories: embedding-based approaches and finetune-based approaches.
\par Embedding-based approaches \cite{xie2016image, mousselly2018multimodal, pezeshkpour2018embedding} can be called multimodal knowledge graph embedding (MMKGE) as well. It follows the paradigm of general KGE and represent each entity and relation with several embeddings. To leverage the multimodal information, these approaches extract multimodal features with pre-trained models like VGG \cite{simonyan2014very} and GloVe \cite{pennington2014glove}, then the multimodal features would be fused into the multimodal embeddings of entities. These methods are backward in extracting multimodal information, rely on a lot of manual design and have 
poor ability to represent the extracted modal information.
\par Finetune-based approaches \cite{yao2019kg,kim2020multi, wang2021structure} would employ pre-trained models like BERT \cite{devlin2018bert} to score the triples directly instead of training entity and relation embeddings. KG-BERT \cite{yao2019kg} extend the triples into text sequences as inputs of BERT and then finetune BERT with the triple classification task. MTL-KGC \cite{kim2020multi} is a multi-task version of KG-BERT. StAR \cite{wang2021structure} apply siamese-style textual encoder to speed up the inference stage. These approaches would are slower and less accurate than the traditional methods of 
inference due to the rank-based evaluation of KGs.

\section{Definition}
\begin{figure*}[h]
  \centering
  \includegraphics[width=0.95\linewidth]{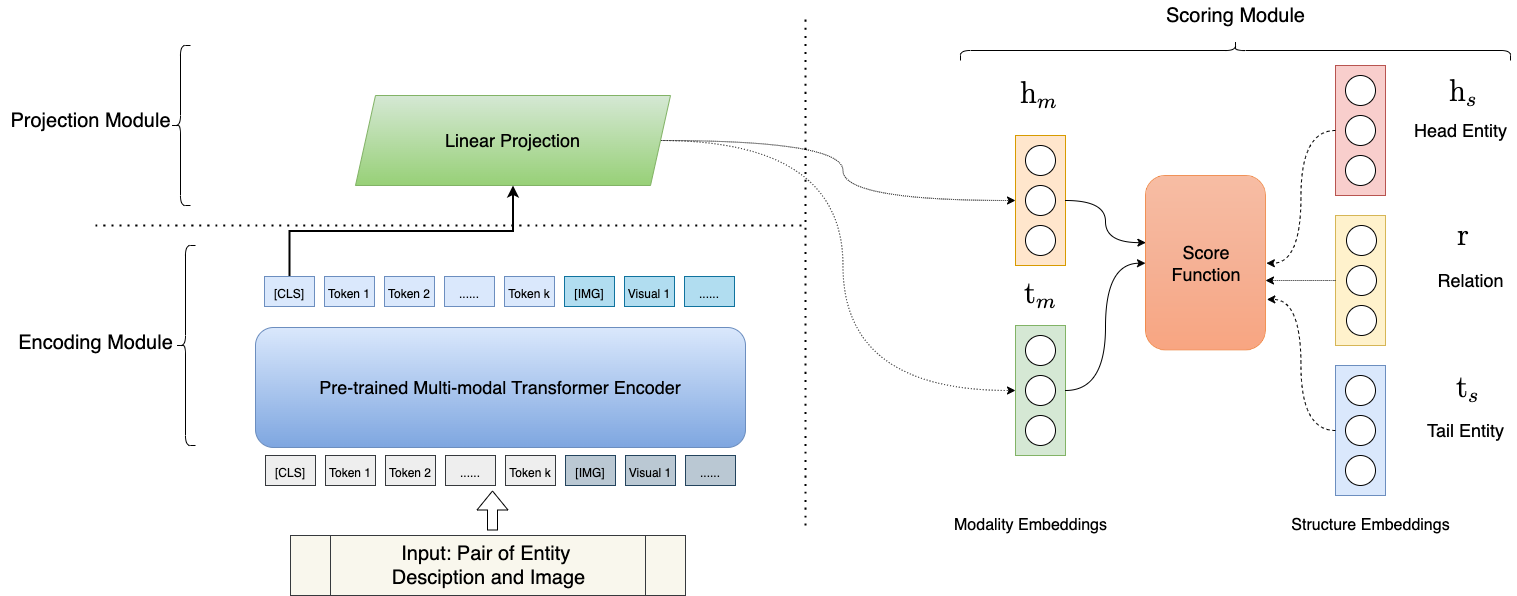}
  \caption{\label{img:model}Model architecture of VBKGE, our model consists of three modules: encoding module, projection module, and scoring module.}
  
\end{figure*}
A MMKG can be denoted as $
\mathcal{G}_{M}=(\mathcal{E},\mathcal{R},\mathcal{I},\mathcal{D}, \mathcal{T})
$, where $\mathcal{E}$ is the entity set,$\mathcal{R}$ is the relation set, $\mathcal{I}$ is the image set, $\mathcal{D}$ is the text description set, $\mathcal{T}=\{(h, r, t) \mid h, t \in \mathcal{E}, r \in \mathcal{R}\}$ is the triple set where $h, r, t$ are the head entity, relation, and tail entitiy of a triple. For each entity $e_i\in\mathcal{E}$, it has a textual description $d_i\in\mathcal{D}$. Each textual description $d_i$ consists of several words, denoted as $d_i=(w_1,w_2,\dots, w_n)$, where $w_j\in \mathcal{V}
$ and $\mathcal{V}$ is the vocabulary of the MMKG $\mathcal{G}_{M}$. Besides, each entity $e_i$ might have 0 to any numbers of images in $\mathcal{I}$, the images of $e_i$ is denoted as set $I_i$.
\par We denote $
\mathbf{e}_s$ and $\mathbf{e}_m$ as the structural embedding and multimodal embedding for an entity $
e$, respectively. Therefore, the entity $
e$  can be represented by two embedding vectors $\mathbf{e}_s, \mathbf{e}_m$. Besides, we denote $\mathbf{r}$ as the embedding of relation $r$.

\section{Method}

The model architecture of our model VBKGE is shown in Figure \ref{img:model}. VBKGE has three modules: encoding module, projection module, and scoring module. The detailed design of each module would be shown below.

\subsection{Encoding Module}
The main role of the encoding module is to encode entities based on their textual and image information. In this paper, we use VisualBERT \cite{li2019visualbert}, a pre-trained multimodal transformer \cite{vaswani2017attention} model as our multimodal encoder. VisualBERT learned the knowledge about how to achieve modal alignment and fusion, which is implicitly stored in the parameters. Thus, we could use pre-trained VisualBERT to obtain deeply integrated multimodal features.
\par For each entity $e_i\in\mathcal{E}$, the textual description of $e_i$ is $d_i=(w_1,w_2,\dots, w_n)$ and the image set of $e_i$ is $I_i$. We use VGG \cite{simonyan2014very} to extract the visual feature and process the images of $e_i$ into several visual tokens $(v_1,
v_2,\dots,v_m)$. The inputs $\hat d_{i}$ of VisualBERT model includes both word tokens and visual tokens:
\begin{equation}
    \hat d_i =([\mathbf{CLS}], w_1,\dots,w_n, [\mathbf{SEP}], v_1, \dots, v_m, [\mathbf{SEP}])
\end{equation}
We add several special tokens like [CLS] and [SEP] following the original VisualBERT paper. The output of VisualBERT is:
\begin{equation}
    \mathbf{VisualBERT}(\hat d_i)=(\mathbf{h}_{\mathbf{CLS}}, \mathbf{h}_{w1}, \ldots, \mathbf{h}_{\mathbf{SEP}}, \mathbf{h}_{v1}, \ldots, \mathbf{h}_{\mathbf{SEP}})
\end{equation}
The hidden state of [CLS] is employed as the initial multimodal feature of each entity $e_i$, denoted as $\mathbf{h}_i$.

\subsection{Projection Module}
The main function of the projection module is to project the modal features of entities into the same representation space of structural embeddings. As the modal features and structured embeddings are heterogeneous, they could not participate in triple scoring together. Thus, we apply a projection matrix $\mathbf{W}$ and obtain the multimodal embedding $\mathbf{e}_{im}$ of each entity $e_i$ by linear projection:
\begin{equation}
    \mathbf{e}_{im}=\mathbf{W}\mathbf{h}_{\mathbf{CLS}}
\end{equation}

\subsection{Scoring Module}
Scoring module would define a score function and estimate the plausibility of each triple. The general priciple of score function is to give higher scores for positive triples and lower scores for negative ones.
\par For each triple $(h, r, t)\in \mathcal{T}$, the score function $\mathcal{F}$ of VBKGE can be divided into five different parts:
\begin{equation}
    \mathcal{F}_{ss}=f(\mathbf{h}_s, \mathbf{r},\mathbf{t}_s)
\end{equation}
\begin{equation}
    \mathcal{F}_{mm}=f(\mathbf{h}_m, \mathbf{r},\mathbf{t}_m)
\end{equation}
\begin{equation}
    \mathcal{F}_{sm}=f(\mathbf{h}_s, \mathbf{r},\mathbf{t}_m)
\end{equation}
\begin{equation}
    \mathcal{F}_{ms}=f(\mathbf{h}_m, \mathbf{r},\mathbf{t}_s)
\end{equation}
\begin{equation}
    \mathcal{F}_{all}=f(\mathbf{h}_s+\mathbf{h}_m, \mathbf{r},\mathbf{t}_s+\mathbf{t}_m)
\end{equation}
\begin{equation}
    \mathcal{F}(h, r, t)=\mathcal{F}_{ss}+\mathcal{F}_{mm}+\mathcal{F}_{sm}+\mathcal{F}_{ms}+\mathcal{F}_{all}
\end{equation}
\par In VBKGE, we apply TransE as function $f$, which can be denoted as:
\begin{equation}
    f(\mathbf{h},\mathbf{r},\mathbf{t})=-||\mathbf{h}+\mathbf{r}-\mathbf{t}||_p
\end{equation}
\par In our scoring function $\mathcal{F}$, the multimodal embeddings and structure embeddings of entities could interact fully with each other as we define multiple score functions. The five score functions could be divided into two parts, unimodal scores, and multimodal scores. Unimodal scores are calculated by one kind of embeddings (structural or multimodal) while multimodal scores need both. Thus, the overall score function $\mathcal{F}$ can be expressed in another way:
\begin{equation}
    \mathcal{F}_{unimodal}=\mathcal{F}_{ss}+\mathcal{F}_{mm}
\end{equation}
\begin{equation}
    \mathcal{F}_{multimodal}=\mathcal{F}_{sm}+\mathcal{F}_{ms}+\mathcal{F}_{all}
\end{equation}
\begin{equation}
    \mathcal{F}(h, r, t)=\mathcal{F}_{unimodal}+\mathcal{F}_{multimodal}
\end{equation}

\subsection{Training Objective And Negative Sampling Strategy}
\subsubsection{Contrastive Training Objective}
As a general paradigm, KGE models would give higher scores for positive triples and lower scores for negative ones. We first generate a negative triple set by randomly replacing the head or tail entity in each positive triple, which is called negative sampling. The negative triple set can be denoted as:
\begin{equation}
    \begin{aligned}
        \mathcal{T}^{\prime}=\left\{\left(h, r, t^{\prime}\right) \mid t^{\prime} \in \mathcal{E} \wedge\left(h, r, t^{\prime}\right) \notin \mathcal{T}\right\} \\ \cup \left\{\left(h^{\prime}, r, t\right) \mid h^{\prime} \in \mathcal{E} \wedge\left(h, r, t^{\prime}\right) \notin \mathcal{T}\right\}
    \end{aligned}
\end{equation}
where $(h,r,t)$ is a positive triple. We genearate $k$ negative samples for each postive triple.
    
Therefore, we apply a margin-rank loss for postive-negative contrast during training:
\begin{equation}
    \mathcal{L}=\sum_{(h, r, t)\in \mathcal{T}}\max \left( \gamma-\mathcal{F}(h, r, t) +\frac{1}{k}\sum_{(h'_i, r'_i, t'_i) \in \mathcal{T}^\mathcal{\prime}}\mathcal{F}(h'_i, r'_i, t'_i), 0\right)
\end{equation}
where $\gamma$ is the margin.
\subsubsection{Twins Negaitve Sampling}
In this paper, we also propose a negative sampling called twins for better MKGE model training. 
\par Traditional negative sampling in KGE is entity-level which would replace the whole head or tail entity to generate a negative triple. It works in general KGE as general KGE usually defines only one embedding for each entity. However, in MKGE models, each entity would have multiple embeddings such as structural and multimodal embeddings. In the multimodal scenario, entity-level negative sampling would replace all the embeddings of the selected entity, which implicitly assumes that different embeddings of this entity have
been aligned. Therefore, such an assumption might be too strong for model training.
\par Hence, we propose a more fine-grained negative sampling strategy called twins to solve the problem. Twins negative sampling employs different negative sampling strategies for unimodal and multimodal parts of the model. That's why we call it twins. With twin negative sampling, the model could not only learn to discriminate the plausibility of triples (just like traditional negative sampling) but also align the different embeddings for each entity.
\par As for the unimodal scores, twins employs just normal negative sampling. The head or tail entity $e$ in the positive triple is randomly replaced by another entity $e'$. For multimodal scores, however, we only sample negative multimodal features for the replaced entity. We still denote the sampled entity as $e'$, but the structural and multimodal embeddings of $e'$ are $\mathbf{e}_s, \mathbf{e}'_m$ while they are $\mathbf{e}'_s, \mathbf{e}'_m$ in normal negative sampling. By 
contrasting with the negative modal features, the model can further align the two kinds of embeddings. With such a fine-grained and modal-level negative sampling strategy for multimodal scores, the model could learn to align the embeddings for each entity during training.
\section{Experiments}
\begin{table}[h]
    \caption{\label{tab:dataset}Statistical information of datasets.}
    \centering
   \resizebox{\columnwidth}{!}{
\begin{tabular}{cccccc}
\toprule
Dataset           & Entities   & Relations   & Train    & Valid   & Test   \\ \hline
WN9           & 6555  & 9 & 11741  & 1337  & 1319  \\ 
FB15K-237  & 14541 & 237 & 272115 & 17535 & 20466 \\ \bottomrule
\end{tabular}
}
\end{table}
In this section, we will report the experiment details including datasets, evaluation protocols, parameter settings, and the results. In addition to the conventional link prediction experiments, we have three exploratory questions:
\begin{enumerate}
    \item \textbf{Question 1} (Q1): Does twins negative sampling works?
    \item \textbf{Question 2} (Q2): Does our methods inference faster than finetune-based approaches?
    \item \textbf{Question 3} (Q3): Whether the design of each part of the model is valid?
\end{enumerate}
\par Following the three questions, we would expolore more about VBKGC next.
\subsection{Datasets}
In our experiments, we employ two public benchmarks WN9 \cite{xie2016image} and FB15K-237 \cite{yao2019kg}. The image resources of FB15K-237 is collected from \cite{liu2019mmkg}. The detailed information about the datasets is shown in Table \ref{tab:dataset}.

\subsection{Evaluation Protocols}
Following classic KG research, we apply link prediction task and rank-based evaluation protocol. Given a correct triple, we rank it against all candidate triples with their scores. Both head and tail entity prediction would be applied in link prediction. The whole entity set $\mathcal{E}$ would be the candidate entity set.
\par We use mean reciprocal rank (MRR) and Hit@K (K=1,3,10) as evaluation metrics. They can be denoted as:
\begin{equation}
    \mathrm{MRR}=\frac{1}{2|\mathcal{T}_{test}|}\sum_{t\in\mathcal{T}_{test}}(\frac{1}{rank_{th}} +  \frac{1}{rank_{tt}})
\end{equation}
\begin{equation}
    \mathrm{Hit@K}=\frac{1}{2|\mathcal{T}_{test}|}\sum_{t\in\mathcal{T}_{test}}
    \mathbf{1}(rank_{th}\le K) +  \mathbf{1}(rank_{tt} \le K)
\end{equation}
where $\mathcal{T}_{test}$ is the test triple set and $rank_{th}, rank_{tt}$ are predicted ranks of head/tail entity prediction for each test triple $t$.
\par Besides, all the metrics are in the filter setting \cite{transe}. It would remove the candidate triples which have already appeared in train and valid data. 

\subsection{Experiment Settings}

For experiments, we set both structural embedding and multimodal embedding size $d_e=128$ for each model. The dimension of multimodal features captured by the pre-trained VisualBERT model is $d_m=768$. For those entities which have no image, we employ Xavier initialization \cite{DBLP:journals/jmlr/GlorotB10} for their visual features. We set the amount of negative sample $k=16$.
\begin{table*}[]

\caption{Expeirment results of the link prediction task. The baselines marked with * are our reproduction based on the original paper. The best results in each metric are bold and the second-best results are underlined. Some results of baselines that are hard for reproduction and have no results in origin paper are marked as -.}
\label{tab:lp}
\begin{tabular}{ccccccccc}

\toprule
\multirow{2}{*}{Method} & \multicolumn{4}{c}{WN9}                                           & \multicolumn{4}{c}{FB15K-237}                                     \\
                        & MRR            & Hit@10         & Hit@3          & Hit@1          & MRR            & Hit@10         & Hit@3          & Hit@1          \\ \hline
TransE*                  & {\ul 0.766}    & 0.912          & 0.885          & {\ul 0.641}    & 0.261          & 0.437          & 0.291          & 0.173          \\
IKRL*                    & 0.433              & 0.938          & 0.849              & 0.011              & 0.268          & 0.449          & 0.301          & 0.177          \\
TransAE                 & -              & {\ul 0.942}    & -              & -              & -              & -              & -              & -              \\
MTKRL*                  & 0.354          & \textbf{0.948} & 0.651          & 0.112          & -              & -              & -              & -              \\
KG-BERT                 & -              & -              & -              & -              & 0.237          & 0.427          & 0.260          & 0.144          \\
StAR(base)              & -              & -              & -              & -              & 0.296          & \textbf{0.482} & 0.322          & 0.205          \\\hline
VBKGC+Normal            & 0.749          & 0.919          & {\ul 0.901}    & 0.592          & {\ul 0.299}    & 0.477          & {\ul 0.331}    & {\ul 0.210}    \\
VBKGC+Twins              & \textbf{0.857} & 0.922          & \textbf{0.904} & \textbf{0.803} & \textbf{0.301} & {\ul 0.478}    & \textbf{0.332} & \textbf{0.213} \\ \bottomrule
\end{tabular}
\end{table*}

During training, we divide each dataset into mini-batches and apply TransE \cite{transe} as base score functions $f$. We use default Adam optimizer for optimization and tune the hyper-parameters of our model with grid search. The number of batches is tuned in $\{100, 400\}$ The margin $\gamma$ is tuned in $\{4.0, 6.0, 8.0, 10.0\}$ and learning rate is tuned in $\{2e-5, 1e-4, 5e-4, 1e-3\}$. The parameter settings are based on existing research findings \cite{xie2016image,han2018openke}. All the experiments are conducted on Nvidia GeForce 3090 GPUs. 
\par As for baselines, we employ several embedding-based methods ( TransE \cite{transe}, IKRL \cite{xie2016image}, TransAE\cite{wang2019multimodal}, MTKRL \cite{mousselly2018multimodal}, all of them apply TransE as score function) and finetune-based approaches (KG-BERT \cite{yao2019kg}, StAR \cite{wang2021structure}). For fair comparisions, we use the same embeddings dim for embedding-based approaches ($d=128$) and reproduce some of the baselines with the same hyperparamters as original papers.

\subsection{Link Prediction Results}

The main experiment results on the link prediction task are shown in Table \ref{tab:lp}. We reproduce some classic baselines on the datasets and those results are marked with *. Some results of baselines that are hard for reproduction and have no results in origin paper are marked as -. We could visualize that our method VBKGC could perform better than baselines on most of the metrics, except Hit@10 on WN9.
\par Besides, the twins negative sampling could behave better than normal negative sampling in the multimodal scenario. The performance gains it brings are particularly noticeable in the WN9 dataset. Though twins negative sampling gets limited improvement on the FB15K237 dataset, we would make a further exploration in the analysis of Q1.
\par Another surprising conclusion that can be deduced from the experimental results is that VBKGC with twins negative sampling could perform precise reasoning. Compared with other embedding-based baselines (\cite{transe, xie2016image, mousselly2018multimodal}), VBKGC with twins could achieve outstanding improvement on Hit@1 and MRR metrics. VBKGC with twins obtains nearly 25\% (from 0.641 to 0.803) and 20\% (from 0.177 to 0.213) on Hit@1 with WN9 and FB15K-237 respectively. Compared with finetune-based approaches, the improvements brought about by our model are also clearly perceptible.

\subsection{Analysis of Q1: Negative Sampling}
\par We could conclude that twins negative sampling could improve the link prediction performance from the previous results. In this section, we would dive deeper into the negative sampling in MMKGC and try to answer Q1.
\par We employ several state-of-the-art negative sampling methods for general KGE (NSCaching \cite{zhang2019nscaching}, NoSampling\cite{li2021efficient}) for comparison. We implement the VBKGC model based on their open-source code and conduct experiments with the same embedding dimension $d_e=128$ on the WN9 dataset. We tuned the hyperparameters according to the original paper and the results are shown in Table \ref{tab:ns}.
\begin{table}[h]
\caption{Expeirment results of different negative sampling methods on WN9. The best results in each metric are bold and the second best results are underlined.}
\label{tab:ns}
\begin{tabular}{ccccc}
\toprule
          & MRR            & Hit@10         & Hit@3          & Hit@1          \\ \hline
Normal    & {\ul 0.749}    & {\ul 0.919}    & {\ul 0.901}    & 0.592          \\
NSCaching & 0.725          & 0.868          & 0.804          & {\ul 0.630}    \\
NoSampling    & 0.426          & 0.662          & 0.492          & 0.306          \\
Twins     & \textbf{0.857} & \textbf{0.922} & \textbf{0.904} & \textbf{0.803}\\
\bottomrule
\end{tabular}
\end{table}

\par We could found that negative sampling methods for general KGE might not be acclimatized for the multimodal scenario as we make our best to tune hyperparameters for better results. They even get a marked regression on some metrics. The performance of twins negative sampling on the WN9 dataset exceeds the existing baselines in all aspects. It could align structural and multimodal embeddings to achieve better performance in the multimodal scenario.
\par
\begin{figure}[h]
  \centering
  \includegraphics[width=0.95\linewidth]{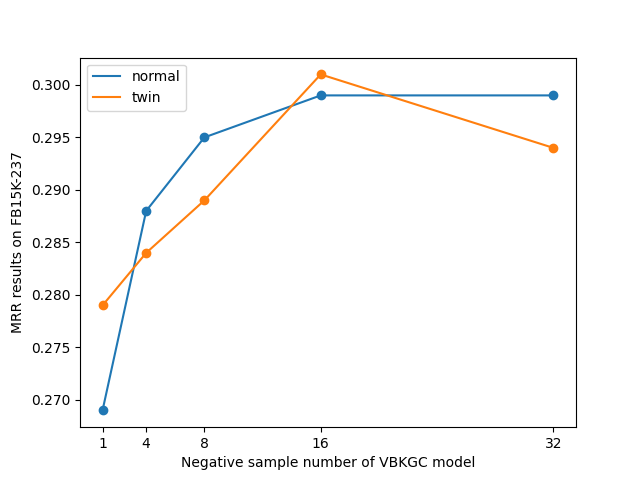}
  \caption{\label{img:ns}Link prediction performance (MRR) on FB15K-237 with different $k$ (numbers of negative samples).}
  
\end{figure}

 Besides, as twins negative sampling gets limited improvement on FB15K237 dataset, we make a further exploration about this problem. We trained several VBKGC models with different amounts of negative samples for both normal and twins negative sampling on FB15K-237 dataset. The evaluation results are plotted as a line graph (Figure \ref{img:ns}).
\par We could observe that when $k=1$,  twins negative sampling could perform better than normal negative sampling. But when $k$ increases, the impact of the twins negative sampling on the VBKGC model seems to be less significant than normal negative sampling. Existing research shows that increasing the number of negative samples is also an effective means \cite{rotate} to improve the model performance. It might be a better choice than twins negative sampling which aligns different embeddings in the FB15K-237 dataset. Nonetheless, it is still a good design for the multimodal scenario as it brings effective enhancement to the WN9 dataset, only it has yet more exploration.

\subsection{Analysis of Q2: Inference Speed}
\par
\begin{figure}[h]
  \centering
  \includegraphics[width=0.95\linewidth]{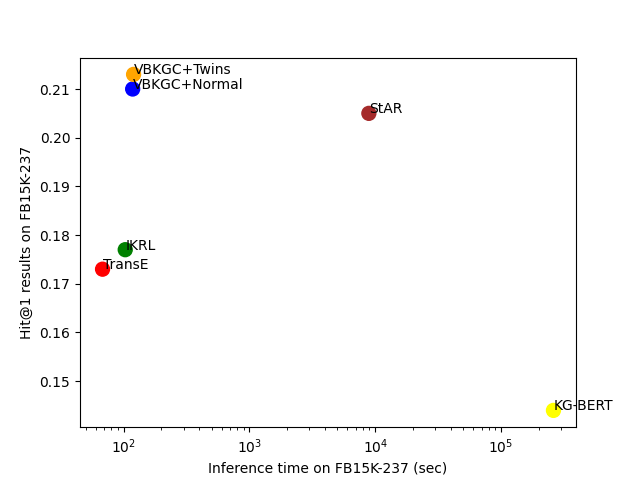}
  \caption{\label{img:time}Inference time and MRR results of different models on FB15K-237 dataset.
  }
\end{figure}

 To answer Q2, we employ several MMKGC models and measure the overall time they need to infer on the FB15K-237 test data. We plot the results of the measurements as a scatter plot shown in Figure \ref{img:time}.
 
\par From the figure we can learn that VBKGC could inference fast like many other embedding-based approaches (TransE \cite{transe}, IKRL \cite{xie2016image}). Compare with finetune-based methods (KG-BERT \cite{yao2019kg}, StAR \cite{wang2021structure}), VBKGC could achieve both better link prediction performance and faster inference speed. Besides, VBKGC would perform a bit slower than other embedding-based methods as VBKGC employs a more complex score function than baselines.

\subsection{Analysis of Q3: Ablation Study}

\par To further prove the effects of different modules in VBKGC, we conduct the ablation study with six different settings of experiments (S1-S6). The settings of S1 to S6 are as follows: (1)(Encoding Module) S1 refers to the model with random multimodal embeddings without VisualBERT to verify the quality of multimodal features captured by VisualBERT. (2) (Scoring Module) S2 to S5 apply just several parts of the overall score function $\mathcal{F}(h, r, t)$. S2 only employs $\mathcal{F}_{ss}$ as a score function. S3 employs $\mathcal{F}_{all}$. S4 employs the unimodal scores $\mathcal{F}_{ss}+\mathcal{F}_{mm}$. S5 employs the multimodal scores $\mathcal{F}_{sm}+\mathcal{F}_{ms}$. (3) (Training Objective) S6 sets $k=1$ and samples only 1 negative triple during training.

\begin{table}[h!]
\caption{Expeirment results of ablation study on FB15K-237. The exact meaning of S1 to S6 could be found in the main text.}
\label{tab:ab}
\begin{tabular}{ccccc}
\toprule
      & MRR   & Hit@10 & Hit@3 & Hit@1 \\ \hline
VBKGC & 0.299 & 0.477  & 0.331 & 0.210 \\
S1: w/o VisualBERT    & 0.226 & 0.409  & 0.261 & 0.132 \\
S2: only $\mathcal{F}_{ss}$    & 0.147 & 0.242  & 0.153 & 0.097 \\
S3: only $\mathcal{F}_{all}$    & 0.261 & 0.435  & 0.293 & 0.173 \\
S4: only $\mathcal{F}_{ss}+ \mathcal{F}_{mm} $   & 0.251 & 0.407  & 0.273 & 0.174 \\
S5: only $\mathcal{F}_{sm}+ \mathcal{F}_{ms} $    & 0.285 & 0.464  & 0.317 & 0.195 \\
S6: only 1 negative   & 0.269 & 0.436  & 0.297 & 0.184 \\

\bottomrule
\end{tabular}
\end{table}

\par The results of the ablation study are shown in Table \ref{tab:ab}. We could find that all of S1 to S6 get worse performance than the full model (VBKGC with normal negative sampling). Thus, the design of each part of our model VBKGC is necessary to get better performance.

\section{Conclusion}

\par In this paper, we present an embedding-based model VBKGC for MMKGC, which employs VisualBERT as a multimodal feature encoder. We achieve co-design of both model and negative sampling by proposing twins negative sampling to align different embeddings for the multimodal scenario. VisualBERT extracts deeply fused multimodal information for better link prediction, which is free of finetuning and makes VBKGC inference fast and precise. Extensive experiment results on two datasets and link prediction tasks with three further explorations demonstrate the effectiveness of VBKGC.
\par In the future, we plan to 1) explore more effective ways to leverage multimodal information in knowledge graphs to benefit more kinds of in-KG and out-KG tasks; 2) find a more expressive architecture and try to pre-train KGs on it to capture the deep knowledge in KGs; 3) borrowing solutions from other multimodal machine learning tasks for multimodal knowledge graphs.


\bibliographystyle{ACM-Reference-Format}
\bibliography{sample-base}


\clearpage
\appendix

\section{Reproducibility}
\subsection{Implemention Details}
We implement our methods based on the open-source library OpenKE. We utilize Pytorch to conduct experiments with one NVIDIA RTX 3090 GPU. Other information about parameter selection is mentioned in the previous section 5.3.
\subsection{Datasets}
The FB15k-237 dataset is available \href{https://github.com/yao8839836/kg-bert/tree/master/data}{here}. The WN9 datasets is available in \href{https://github.com/UKPLab/starsem18-multimodalKB}{here} . We use the images of FB15K-237 released \href{https://github.com/mniepert/mmkb}{here}.
\subsection{Optimal parameters}
The optimal hyper-parameters of our model VBKGC with both negative sampling methods on WN9 dataset is:
\begin{enumerate}
    \item number of batches: 100
    \item margin $\lambda$: 8
    \item learning rate: 2e-5
\end{enumerate}

The optimal hyper-parameters on FB15K-237 dataset is:
\begin{enumerate}
    \item number of batches: 400
    \item margin $\lambda$: 6
    \item learning rate: 2e-5
\end{enumerate}

\end{document}